\documentclass{article}

\usepackage{PRIMEarxiv}
\usepackage{amsthm}  

\newtheorem{theorem}{Theorem}[section]

\newtheorem{proposition}[theorem]{Proposition}

\theoremstyle{remark}
\newtheorem{remark}{Remark}[section]\usepackage[ruled,vlined]{algorithm2e}

\usepackage[utf8]{inputenc} 
\usepackage[T1]{fontenc}    
\usepackage{hyperref}       
\usepackage{url}            
\usepackage{booktabs}       
\usepackage{amsfonts}       
\usepackage{nicefrac}       
\usepackage{microtype}      
\usepackage{lipsum}
\usepackage{fancyhdr}       
\usepackage{graphicx}       
\usepackage{paralist}  
\usepackage{bm}  
\usepackage{placeins}

\usepackage{amsmath} 
\usepackage{multirow} 
\usepackage{tabularx}
\usepackage{booktabs}
\usepackage{float}

\usepackage{threeparttablex} 
\usepackage{pifont}

\newcolumntype{Y}{>{\centering\arraybackslash}X}
\usepackage[numbers]{natbib} 

\graphicspath{{media/}}     

\title{\textbf{\texttt{AdaFair-MARL}}: Enforcing Adaptive Fairness Constraints in Multi-Agent Reinforcement Learning
\thanks{\textit{\underline{Citation}}: 
\textbf{Promise Osaine Ekpo, Brian La, Thomas Wiener, Saesha Agarwal, Arshia Agrawal, Gonzalo Gonzalez-Pumariega, Lekan P.\ Molu, Angelique Taylor. Skill-Aligned Fairness in Multi-Agent Learning for Collaboration in Healthcare. Pages.... DOI:000000/11111.}} 
}

\author{
  Promise Ekpo \\
  Cornell Tech, NY, USA \\
  \texttt{poe6@cornell.edu} \\
  \And
  Saesha Argawal \\
  Cornell University, NY, USA \\
  \texttt{sa2388@cornell.edu} \\
  \And
  Felix Grimm \\
  Cornell University, NY, USA \\
  \texttt{fjg45@cornell.edu} \\
  \And
  Jiachang Liur \\
  Cornell Tech, NY, USA \\
  \texttt{amt298@cornell.edu}
  \And
  Lekan Molu \\
  Bala-Cynwyd, PA, USA \\
  \texttt{lekanmolu@molux-labs.com} \\
  \And
  Angelique Taylor \\
  Cornell Tech, NY, USA \\
  \texttt{amt298@cornell.edu}
}

\begin{document}
\maketitle

\begin{abstract}

Fair workload enforcement in heterogeneous multi-agent systems that pursue shared objectives remains challenging. Fixed fairness penalties often introduce inefficiencies, training instability, and conflicting agent incentives.
Reward-shaping approaches in fair Multi-Agent Reinforcement Learning (MARL) typically incorporate fairness through heuristic penalties or scalar reward modifications and often rely on post-hoc evaluation. However, these methods do not guarantee that a desired fairness level will be satisfied.
To address this limitation, we propose the Adaptive Fairness Multi-Agent Reinforcement Learning (\textbf{\texttt{AdaFair-MARL}}) framework, which formulates workload fairness as an explicit constraint so that agents maintain balanced contributions while optimizing team performance.
We present \textbf{\texttt{AdaFair-MARL}}, a constrained cooperative 
MARL framework whose core algorithmic component is a primal--dual 
update that enforces workload fairness via adaptive Lagrange multiplier 
updates. Grounding the framework in a cooperative Markov game, we 
derive the fairness constraint from Jain's Fairness Index~(JFI) 
geometry and show that the resulting feasible set admits a second-order 
cone representation, enabling principled Lagrangian dual-ascent updates 
without manual penalty tuning.
Experiments in a simulated hospital coordination environment (MARLHospital) demonstrate the effectiveness of \textbf{\texttt{AdaFair-MARL}} compared to reward-shaping and fixed-penalty fairness methods, improving workload balance while maintaining team performance. We found that \textbf{\texttt{AdaFair-MARL}} achieves nearly perfect constraint satisfaction (0.99–1.00) while significantly improving workload fairness compared to fixed-penalty baselines.

\end{abstract}


\section{Introduction}
\label{intro}

Fair multi-agent systems (MAS) play a crucial role in ensuring efficient task allocation and equitable workload distribution across transportation,
resource allocation, and healthcare~\cite{cassandras2025control,%
dong2025task,yuan_multi-agent_2021,ekpo_skill-aligned_2025,%
ekpogeneralized}.
Neglecting fairness in MAS can lead to unbalanced agent workflows, leading to overworked agents, inefficient workload distribution, and poor coordination. 
Relevant to our work, empirical studies in hospital settings show that uneven workloads raise cognitive and temporal demands by 25--40\% and delay critical actions, making fairness a quantitative determinant of patient
safety~\cite{taylor_rapidly_2025,tanjim_help_2025,tanjim_human-robot_2025}.

Our work spans three bodies of literature, including (i) fairness in MARL and (ii) MAS, and (iii) constrained MARL.
Fairness in MARL incorporates fairness metrics into learning objectives through intrinsic rewards or extrinsic signal modifications.
Inequality aversion techniques serve as an intrinsic reward based on pairwise return differences \cite{hughes_inequity_2018}, while methods such as the Gini Index act as a social welfare function \cite{siddique_fairness_nodate}, and Fairness Aware Reinforcement Learning via Proximal Policy Optimization augments demographic-parity penalties \cite{malfa_fairness_2025}.
Similar formulations appear in network and traffic scheduling, embeding fairness objective functions with soft incentives rather than constraints \cite{yuan_multi-agent_2021,fang_fairness-aware_2024}.
In the fairness in MAS literature, proportional fairness, max-min fairness, and throughput fairness have been studied across networked MAS~\cite{khan_fairness_2016,monteiro_fairness_2025}.
Constrained MARL methods such as Constrained Policy Optimization~\cite{achiam_constrained_2017} and first-order constrained optimization~\cite{zhang_first_2020} focus on trajectory-based or per-agent safety constraints, neglecting joint agent fairness constraints.

Despite significant efforts to constrain fairness in MAS and MARL, several gaps remain.
In contrast to prior fairness in MARL research, we treat fairness as an explicit constraint enforced through adaptive dual ascent.
Building on the constrained MARL literature, our constraints couple all agents’ workloads into a shared
condition. 
Unlike work on fairness in MAS, our work uses Jain's Fairness Index
(JFI)~\cite{jain_quantitative_1998} in MARL as a hard constraint enforced during policy optimization, a structure not exploited in MAS fairness formulations.

\begin{figure*}[t]
  \centering
  \includegraphics[width=\textwidth]{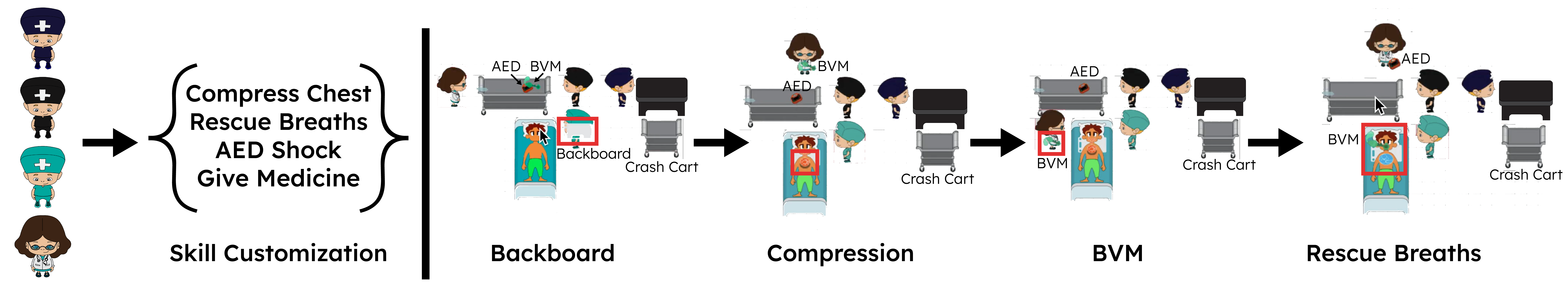}
  \caption{\textbf{MARLHospital \cite{ekpo_skill-aligned_2025}.} A diagnostic MARL benchmark environment for heterogeneous teams with multidimensional fairness metrics. 
With two key features: (1) skill heterogeneity requiring task-agent skill matching, (2) sequential dependencies where skill-task misalignment makes delays compound, 
Agents pick up the backboard,
place it under the patient, perform multiple chest compressions, retrieve the Bag-Valve-Mask, and deliver rescue
breaths.
} 
  \label{fig:teaser}
\end{figure*}

To address these gaps, we introduce the Adaptive Fairness Multi-Agent Reinforcement Learning (\textbf{\texttt{AdaFair-MARL}}) framework, which addresses three challenges:
(i)~measuring workload balance via JFI~\cite{jain_quantitative_1998}---a quasi-concave, scale-invariant fairness
metric widely used in networking \cite{wei_equinox_2025, islam_equity-aware_nodate} and scheduling \cite{zhou_fairness-aware_nodate}; (ii)~enforcing the constraint $\mathrm{JFI}(\mathbf{w})\ge\tau$
where $\tau\in(0,1)$ is a learnable fairness threshold; and
(iii)~a dual-ascent update that adaptively enforces the constraint
during learning, eliminating manual penalty tuning. 
This paper makes three contributions: (i) a fairness-constrained cooperative MARL formulation using Jain's fairness index (JFI) expressed as a second-order cone (SOC) constraint, (ii) a primal--dual training procedure that adaptively enforces fairness thresholds during learning, and (iii) empirical evaluation in a heterogeneous healthcare coordination simulator. We evaluate \textbf{\texttt{AdaFair-MARL}} in the MARLHospital environment \cite{ekpo_skill-aligned_2025} (see Fig.~\ref{fig:teaser}), which is well suited to this problem because it combines heterogeneous agent skills with sequential task dependencies that make workload imbalance consequential.
Our findings show that \textbf{\texttt{AdaFair-MARL}} achieves high success rates while maintaining fairness thresholds unattainable by unconstrained baselines. The remainder of this paper is organized as follows: 
Section~\ref{sec:method} presents the \textbf{\texttt{AdaFair-MARL}} framework, Section~\ref{sec:exp} presents experiments,  Section~\ref{sec:results} reports empirical results, and Section~\ref{sec:conclusion} concludes with implications for fair MARL.

\section{Fairness-Constrained Cooperative MARL}
\label{sec:method}


The \textbf{\texttt{AdaFair-MARL}} framework formalizes a cooperative Markov game and derives a Lagrangian dual-ascent update for adaptive fairness constraint enforcement.

\subsection{Problem Formulation}

MARL can be viewed as a special 
case of a stochastic dynamic game, where multiple decision makers 
interact through a shared environment and a coupled objective. We 
formalize this connection using standard MDP notation~\cite{li_confluence_2023}.

\paragraph{Markov Game}
A discounted Markov game is defined as
$\mathcal{M}=(\mathcal{N},\mathcal{S},\{\mathcal{A}_i\}_{i=1}^{n},T,r,\gamma,d)$,
where $n \in \mathbb{N}$ denotes the number of agents,
$\mathcal{N}=\{1,\dots,n\}$ the set of agents,
$\mathcal{S}$ the global state space, and $\mathcal{A}_i$ the local action
space of agent $i$.
The transition kernel
$T:\mathcal{S}\times\prod_i\mathcal{A}_i\!\to\!\Delta(\mathcal{S})$
determines how the next state is sampled, while
$r:\mathcal{S}\times\prod_i\mathcal{A}_i\!\to\!\mathbb{R}$ is the shared
instantaneous reward. $\gamma\!\in\!(0,1)$ is the discount factor and $d$
the initial-state distribution.
At each time step $t \in \{0,1,2,\dots\}$, each agent
$i \in \mathcal{N}$ samples an action $a_{i,t} \in \mathcal{A}_i$ from
its local stochastic policy
$\pi_i(\cdot\mid s_t) : \mathcal{S} \to \Delta(\mathcal{A}_i)$.
The joint action
$\mathbf{a}_t=(a_{1,t},\dots,a_{n,t}) \in \prod_{i=1}^{n}\mathcal{A}_i$
is executed and $s_{t+1}\sim T(s_t,\mathbf{a}_t)$.

\paragraph{Objective}
The common expected discounted return under a stationary joint policy
$\pi=\prod_{i=1}^{n}\pi_i$ is:
\begin{align}
J(\pi)=\mathbb{E}_{s_0\sim d}\!\left[\sum_{t=0}^{\infty}
\gamma^t\,r(s_t,\mathbf{a}_t)\right],
\label{eq:team_return}
\end{align}
where $J(\pi)$ denotes the total expected team reward discounted over an
infinite horizon from $s_0\sim d$.
We consider a fully cooperative setting where $u_i(\pi)\equiv J(\pi)$
for all $i\in\mathcal{N}$, so $J(\pi)$ serves as the common team
objective. Coupling arises solely through the shared fairness constraint.

\subsection{Embedding a Fairness Constraint}

To capture equitable workload distribution, we augment the Markov game
with a shared JFI constraint based on a differentiable fairness index.

\paragraph{Fairness Constraint}
Let $w_{i,t}$ denote agent $i$'s cumulative validated task completions
up to time $t$, with workload vector $\mathbf{w}_t=[w_{1,t},\dots,w_{n,t}]$.
The Jain Fairness Index (JFI) is
\begin{align}
\mathsf{F}(\mathbf{w}_t)=
\frac{\left(\sum_{i=1}^{n} w_{i,t}\right)^2}
{n\sum_{i=1}^{n} w_{i,t}^2}
\in[0,1],
\label{eq:jfi}
\end{align}
where $\mathsf{F}=0$ indicates maximally unfair and $\mathsf{F}=1$ indicates perfectly equal 
workload distribution across all agents.
The fairness constraint is:
\begin{align}
g(s_t)=\tau-\mathsf{F}(\mathbf{w}_t)\le 0, \qquad \tau\in(0,1).
\end{align}

A feasible policy therefore satisfies $\mathsf{F}(\mathbf{w}_t)\ge\tau$.
Because workloads correspond to counts of completed tasks,
we have $\mathbf{w}\in\mathbb{R}_+^n$, where $\mathbb{R}_+^n = \{\mathbf{w} \in 
\mathbb{R}^n \mid w_i \ge 0\ \forall i \in \mathcal{N}\}$ 
denotes the non-negative orthant over all $n$ agents. We also write 
$\mathbf{1}\in\mathbb{R}^n$ for the all-ones vector, so that 
$\mathbf{1}^\top w = \sum_{i=1}^n w_i$ denotes the total workload 
summed across all $n$ agents $i \in \mathcal{N}$. The set 
$\mathbb{R}_+^n\setminus\{\mathbf{0}\}$ excludes the zero vector, at which 
$\mathsf{F}(\mathbf{w}_t)$ is undefined.
$\tau \in (0,1]$ defines the feasible workload set as:

\begin{proposition}
\label{prop:convex}
For any $\tau\in(0,1]$, the set
$\mathcal{C}_\tau=\bigl\{\mathbf{w}\in\mathbb{R}_+^n\setminus\{\mathbf{0}\}\mid\mathsf{F}(\mathbf{w})\ge\tau\bigr\}$
is convex.
\end{proposition}
\begin{proof}
For $\mathbf{w}\in\mathbb{R}_+^n\setminus\{\mathbf{0}\}$,
$\mathsf{F}(\mathbf{w})\ge\tau \iff (\mathbf{1}^\top \mathbf{w})^2\ge n\tau\|\mathbf{w}\|_2^2$.
Since $\mathbf{1}^\top \mathbf{w}\ge0$, taking square roots gives
$\|\mathbf{w}\|_2\le\frac{1}{\sqrt{n\tau}}\mathbf{1}^\top \mathbf{w}$, so:
\[
\mathcal{C}_\tau=\left\{\mathbf{w}\in\mathbb{R}_+^n \setminus \{\mathbf{0}\} \;\middle|\;
\|\mathbf{w}\|_2\le\frac{1}{\sqrt{n\tau}}\mathbf{1}^\top w\right\}.
\]
This is a second-order cone representable set of the form
\[
\|w\|_2 \le t, \qquad
t=\frac{1}{\sqrt{n\tau}}\,\mathbf{1}^\top w,
\]
and second-order norm cones are convex \cite{boyd_convex_2023}.
Hence $\mathcal{C}_\tau$ is convex.
Similar SOC reformulations of Jain’s fairness index constraints
have been used in optimization problems \cite{rezaeinia2023efficiency}.
\end{proof}


\begin{remark}
$\mathsf{F}$ is quasiconcave on $\mathbb{R}_+^n\setminus\{\mathbf{0}\}$; 
the superlevel set $\mathcal{C}_\tau$ is convex for all $\tau\in(0,1]$ 
via the SOC representation 
$\|\mathbf{w}\|_2\le\frac{1}{\sqrt{n\tau}}\mathbf{1}^\top\mathbf{w}$~\cite{boyd_convex_2023}. 
Here, ``second-order cone'' refers to a convex geometric structure 
defined by a Euclidean norm constraint, not to second-order agent dynamics. This is in alignment with prior work\cite{sundar2024parametricsecondorderconerepresentable}, 
\end{remark}

\paragraph{Quasiconcavity of \texorpdfstring{\(\mathsf{F}\)}{F} and quasiconvexity of \texorpdfstring{\(g\)}{g}}
\begin{proposition}[Quasiconcavity of the Jain fairness index]
\label{prop:Ctau_convex_final}
Let $\mathsf{F}(\mathbf{w})=\frac{(\mathbf{1}^\top\mathbf{w})^2}{n\|\mathbf{w}\|_2^2}$, $\mathbf{w}\in\mathbb{R}_+^n\setminus\{\mathbf{0}\}$.
Then $\mathsf{F}$ is quasiconcave on $\mathbb{R}_+^n\setminus\{\mathbf{0}\}$; equivalently, $g(\mathbf{w}):=\tau-\mathsf{F}(\mathbf{w})$ is quasiconvex for any fixed $\tau\in(0,1]$.
\end{proposition}
\begin{proof}
It suffices to show every superlevel set $\mathcal{C}_\beta=\{\mathbf{w}\in\mathbb{R}_+^n\setminus\{\mathbf{0}\}\mid\mathsf{F}(\mathbf{w})\ge\beta\}$ is convex.
\textit{Case 1} ($\beta\le 0$): $\mathsf{F}(\mathbf{w})\ge0$ always, so $\mathcal{C}_\beta=\mathbb{R}_+^n\setminus\{\mathbf{0}\}$, which is convex up to the harmless exclusion of the origin where $\mathsf{F}$ is undefined.
\textit{Case 2} ($\beta>1$): $\mathsf{F}(\mathbf{w})\le1$ always, so no vector satisfies $\mathsf{F}(\mathbf{w})\ge\beta$, giving $\mathcal{C}_\beta=\emptyset$, which is convex.
\textit{Case 3} ($\beta\in(0,1]$): $\mathsf{F}(\mathbf{w})\ge\beta\iff(\mathbf{1}^\top\mathbf{w})^2\ge n\beta\|\mathbf{w}\|_2^2$. Since $\mathbf{1}^\top\mathbf{w}\ge0$, taking square roots preserves the inequality, giving $\|\mathbf{w}\|_2\le\frac{1}{\sqrt{n\beta}}\mathbf{1}^\top\mathbf{w}$, so $\mathcal{C}_\beta=\{\mathbf{w}\in\mathbb{R}_+^n\mid\|\mathbf{w}\|_2\le\frac{1}{\sqrt{n\beta}}\mathbf{1}^\top\mathbf{w}\}$, a second-order cone constraint, hence convex~\cite{boyd_convex_2023}.
Thus every superlevel set of $\mathsf{F}$ is convex, so $\mathsf{F}$ is quasiconcave. For $g$, its $\alpha$-sublevel set $\{\mathbf{w}\mid g(\mathbf{w})\le\alpha\}=\{\mathbf{w}\mid\mathsf{F}(\mathbf{w})\ge\tau-\alpha\}$ is a superlevel set of $\mathsf{F}$, hence convex. So $g$ is quasiconvex.
\end{proof}

\paragraph{Discounted Cumulative Constraint}
Enforcing $g(s_t)\le0$ at every step is intractable. Following the
constrained MDP formulation~\cite{achiam_constrained_2017}, we use:
\begin{align}
\bar{g}(\pi)=\mathbb{E}_{\pi}\!\left[
\sum_{t=0}^{\infty}\gamma^{t}\bigl(\tau-\mathsf{F}(\mathbf{w}_t)\bigr)
\right]\le0.
\label{eq:disc_constraint}
\end{align}
We focus on a single JFI-based workload constraint; multi-metric extensions are left for future work.

\subsection{Convex surrogate and primal--dual interpretation}
Let $\theta\in\Theta\subseteq\mathbb{R}^d$ and define the surrogate problem
\[
\min_{\theta\in\Theta} -J(\pi_\theta)
\quad
\text{s.t.}
\quad
\bar{g}(\pi_\theta)\le0 .
\]

To connect the Lagrangian structure to standard constrained optimization
theory, we analyze a convex surrogate of~\eqref{eq:constrained_marl}
in which $-J(\pi_\theta)$ and $\bar{g}(\pi_\theta)$ are assumed convex
in $\theta$. This is a modeling assumption that simplifies the
theoretical analysis and admits a clean saddle-point characterization;
the case where these assumptions fail (for example, under neural network
policies) is addressed in Remark~\ref{rem:nonconvex}.

\begin{theorem}
\label{thm:convex_surrogate}
Assume $\Theta$ is convex and consider the surrogate case where
$-J(\pi_\theta)$ and $\bar g(\pi_\theta)$ are convex in $\theta$. If Slater's condition holds
($\exists\,\theta_0$ such that $\bar{g}(\pi_{\theta_0})<0$),
then the problem is convex. If the optimal value is finite and attained,
strong duality holds and the Karush–Kuhn–Tucker (KKT) conditions are necessary and sufficient
for global optimality
\cite{boyd_convex_2023}. Moreover the
Lagrangian is:
\[
\mathcal{L}(\theta,\lambda)
=
-J(\pi_\theta)+\lambda\bar{g}(\pi_\theta),
\qquad \lambda\ge0
\]
admits a saddle point $(\theta^\star,\lambda^\star)$.
\end{theorem}

\begin{proof}
Convexity of the objective and constraint implies the problem is a
convex program \cite{boyd_convex_2023}. Slater's condition
implies zero duality gap \cite{boyd_convex_2023} and,
when the optimum is attained, the KKT conditions characterize global
optimality \cite{boyd_convex_2023}. 
\end{proof}

\begin{remark}
\label{rem:nonconvex}
The saddle-point structure motivates the primal--dual update.  
For neural
policies these convexity assumptions typically fail, and gradient
methods are only guaranteed to converge to stationary points
\cite{jin2021nonconvex}. The convex analysis clarifies the geometry of the fairness constraint,
while Section~\ref{sec:results} evaluates constraint satisfaction
empirically under neural policy optimization.
\end{remark}

\paragraph{Optimization Problem}
The fairness-constrained cooperative MARL objective is:
\begin{align}
\max_{\pi}\;J(\pi)\quad\text{s.t.}\quad\bar{g}(\pi)\le0,
\label{eq:constrained_marl}
\end{align}
where $\bar{g}(\pi)$ is defined in~\eqref{eq:disc_constraint}.

\subsection{Lagrangian Relaxation}
\textbf{\texttt{AdaFair-MARL}} uses a nonnegative Lagrange multiplier $\lambda\ge0$ controlling
fairness enforcement strength. The team Lagrangian is given by:
\begin{align}
\mathcal{L}(\pi,\lambda)
&=J(\pi)-\lambda\,\bar{g}(\pi) \notag\\
&=\mathbb{E}\!\left[\sum_{t=0}^{\infty}\gamma^t\Bigl(
r(s_t,\mathbf{a}_t)-\lambda\bigl(\tau-\mathsf{F}(\mathbf{w}_t)\bigr)
\Bigr)\right].
\label{eq:lagrangian}
\end{align}
We parameterize policies as neural networks $\pi_\theta(a\mid s)$, where
$\theta\in\mathbb{R}^d$. An optimal primal--dual pair
$(\theta^\star,\lambda^\star)$ must satisfy the KKT conditions:
\begin{equation}
\begin{aligned}
\nabla_\theta\mathcal{L}(\theta^\star,\lambda^\star)&=0,\quad
\bar{g}(\pi_{\theta^\star})\le0,\\
\lambda^\star&\ge0,\quad
\lambda^\star\bar{g}(\pi_{\theta^\star})=0.
\end{aligned}
\label{eq:kkt}
\end{equation}

These conditions characterize first-order stationary points of the
Lagrangian. In the nonconvex neural-policy setting they are necessary
for local optimality but do not guarantee global optimality under
function approximation \cite{jin2021nonconvex}.

Constraint satisfaction is 
validated empirically in Sec.~\ref{sec:results}.

\begin{algorithm}[t]
\small\SetAlgoLined\LinesNumbered
\caption{\textbf{\texttt{AdaFair-MARL}}: Primal-Dual Policy Iteration}
\label{alg:fair_marl}
\KwIn{%
  $\pi^{(0)}$: initial joint policy;
  $\lambda^{(0)}\!\leftarrow\!0$: initial Lagrange multiplier;
  $\tau\in(0,1)$: fairness threshold (minimum JFI);
  $\eta_\lambda>0$: dual learning rate;
  $\lambda_{\max}>0$: multiplier cap;
  $\gamma\in(0,1)$: discount factor;
  $M$: number of Monte Carlo rollouts per update\;
}
\KwOut{%
  $\pi^{(K)}$: policy satisfying $\mathsf{F}(\mathbf{w}_t)\ge\tau$ with high probability.\;
}
\For{$k=0,1,2,\ldots$}{
  \tcp{Primal: policy update at fixed $\lambda^{(k)}$}
    $\tilde{r}_{\text{ep}} \leftarrow r_{\text{ep}} 
        - \lambda^{(k)}\!\left(\tau - \mathsf{F}(\mathbf{w}_{\text{ep}})\right)$\;
  $\pi^{(k+1)}\approx\arg\max_{\pi}\tilde{J}(\pi;\lambda^{(k)})$\;
  \tcp{Constraint estimate via Monte Carlo}
  $\bar{g}^{(k+1)}\leftarrow\frac{1}{M}
    \sum_{m=1}^{M}\sum_{t=0}^{T_m}\gamma^t
    \!\left(\tau-\mathsf{F}(\mathbf{w}_t^{(m)})\right)$\;
  \tcp{Dual: projected gradient ascent}
  $\lambda^{(k+1)}\leftarrow\mathrm{clip}\!\left(
    \lambda^{(k)}+\eta_\lambda\bar{g}^{(k+1)},\,0,\,\lambda_{\max}\right)$\;
}
\end{algorithm}

\subsection{Primal--Dual MARL Algorithm}

\paragraph{Primal step}
For fixed $\lambda$, agents maximize the shaped reward:
\begin{align}
\tilde{r}_t=r(s_t,\mathbf{a}_t)-\lambda\bigl(\tau-\mathsf{F}(\mathbf{w}_t)\bigr).
\label{eq:shaped_reward}
\end{align}

\paragraph{Dual step}
After each environment step, the constraint violation is estimated
from the current workload vector $\mathbf{w}_t$ and $\lambda$ to a projected gradient ascent with dual learning rate
$\eta_\lambda > 0$:
\begin{align}
\lambda\leftarrow[\lambda+\eta_\lambda\bar{g}(\pi)]_+.
\label{eq:dual_update}
\end{align}
When the constraint is violated, $\lambda$ increases, strengthening the
penalty; when satisfied, $\lambda$ decreases toward zero. This eliminates
manual penalty tuning while allowing the fairness level to be controlled
through $\tau$.

Algorithm~\ref{alg:fair_marl} follows the primal--dual 
formulation of~\cite{achiam_constrained_2017}, extended to cooperative 
MARL with a fairness constraint on agent workloads. The update 
structure is analogous to stochastic saddle-point methods such as 
those analyzed in~\cite{nedic2009subgradient}, although the neural 
policy parameterization introduces nonconvexity and stochastic gradient 
estimation. The constraint violation $\bar{g}(\pi)$ is estimated via 
Monte Carlo rollouts: at each iteration $k$, the current policy 
$\pi^{(k)}$ is rolled out for $M$ episodes and the discounted fairness 
violations $\tau-\mathsf{F}(\mathbf{w}_t)$ are averaged across rollouts 
and timesteps to form $\bar{g}^{(k+1)}$.


\section{Experiments}
\label{sec:exp}
Our experiments answer the questions:
(1) Can adaptive constraint enforcement achieve fairness without degrading task performance?
(2) How does \textbf{\texttt{AdaFair-MARL}} compare to fixed-penalty baselines in satisfying fairness constraints?

\subsection{Setup}
Many MARL environments serve as natural testbeds for experimentation, including Overcooked-AI \cite{carroll_utility_2020}, 
SMACv2 \cite{ellis_smacv2_2023}, and Robotouille \cite{gonzalezpumariega2025robotouilleasynchronousplanningbenchmark}; 
however, they do not support the heterogeneous agent 
skill structure and structured task hierarchies central to our 
fairness formulation. 
Although Overcooked-AI and Robotouille model tasks with temporal dependencies, they assume homogeneous agents and lack support for evaluating different team compositions. 
SMACv2\cite{ellis_smacv2_2023} supports heterogeneous agents, but does not support structured 
task hierarchies required to evaluate workload 
fairness across agents with different skill levels. 
MARLHospital \cite{ekpo_skill-aligned_2025} is well-suited for our experiments because it was designed specifically to address these gaps, providing fully customizable skill levels, and interdependent medical subtasks that model realistic safety-critical teamwork,making it the most appropriate 
benchmark for evaluating fair coordination in 
heterogeneous multi-agent teams (see Fig. \ref{fig:teaser}). 


MARLHospital \cite{ekpo_skill-aligned_2025} models a structured resuscitation procedure with strict sequential dependencies: agents must retrieve and position equipment before each medical subtask can be initiated, and skill-task mismatches cause compounding delays. The rescue breaths (CPR) task used here requires agents to complete a multi-step chain from retrieving the CPR board, placing it on the patient, performing chest compressions, retrieving the BVM, to delivering rescue breaths with each step gated by successful completion of the prior one. This sequential structure, combined with three heterogeneous agents coordinating under partial observability, makes MARLHospital a non-trivial benchmark despite its low-fidelity design, as seen in Figure~\ref{fig:teaser}.
The state $s_t$ encodes symbolic and spatial features; agents have
eight primitives (\texttt{move}, \texttt{pick}, \texttt{place},
\texttt{stack}, \texttt{treat}, \texttt{compress\_chest},
\texttt{give\_rescue\_breaths}, \texttt{noop}).
The shared reward is progress-based:
$r(s_t,\mathbf{a}_t)=H(s_t)-H(s_{t-1})$, where $H(s_t)$ counts
completed task steps.
Workload counters $w_{i,t}$ increment on validated task completions,
excluding \texttt{noop} and \texttt{move}.








\subsection{Baselines}\label{sec:base_algo}

We compare against three baselines using QMIX via the EPyMARL library \cite{papoudakis2021benchmarkingmultiagentdeepreinforcement}.
\textbf{No-fairness QMIX} uses only the task reward with no fairness
penalty ($\lambda=0$).
\textbf{QMIX with fairness penalties} ($\lambda\in\{10,30\}$)~\cite{ekpo_skill-aligned_2025}
integrates workload balance into the reward with fixed penalty weights;
$\lambda$ is constant throughout training and provides no guarantee
that a target fairness level is met. The penalty values $\lambda\in\{10,30\}$ represent a low and 
moderate fixed-penalty range. Since \textbf{\texttt{AdaFair-MARL}} 
eliminates manual penalty tuning entirely, the precise choice 
of baseline $\lambda$ does not affect the main claim: fixed-penalty 
methods provide no constraint satisfaction guarantee regardless 
of $\lambda$, as confirmed by the CSat column in Table~\ref{tab:results}.
\textbf{\texttt{AdaFair-MARL}} enforces
$\mathrm{JFI}(w)\ge\tau$ via dual ascent with dual rate $\eta=0.01$,
update frequency of 1, and $\lambda_{\max}=20$.
The results span $\tau\in\{0.55,0.65,0.75,0.85\}$, for the fairness--efficiency frontier from relaxed to strict constraint. 
We use QMIX\cite{Rashid2020} as the backbone for all baselines because 
Centralized Training with Decentralized Execution (CTDE) algorithm class
achieves the strongest task performance in MARLHospital~\cite{ekpo_skill-aligned_2025} without 
fairness intervention. The workload-balance penalty baseline follows 
the standard reward-shaping approach to workload 
fairness~\cite{ekpo_skill-aligned_2025}.\textbf{\texttt{AdaFair-MARL}} 
is model-agnostic and applicable to any MARL algorithm in
EPyMARL library~\cite{papoudakis2021benchmarkingmultiagentdeepreinforcement}, 
including Independent Proximal Policy Optimization (IPPO)\cite{yu_surprising_2022-1}.


\subsection{Training and Testing Procedure}
Each experiment includes three agents and is conducted across five random seeds, with results averaged across them. The episode length is 50 timesteps, and all models are trained for a total of 1.25M timesteps. Performance is evaluated across three metrics - task success rate, JFI, and constraint satisfaction to jointly assess coordination effectiveness and fairness.
QMIX hyperparameters include hidden dimension 64, learning rate 0.001 
with decay rate 0.95 every 50k steps, Gated Recurrent Unit (GRU) network, batch size 
1024, replay buffer 50k, double Q-learning, target network 
updated per 25 episodes, $\epsilon$ annealed from 1.0 to 0.05 over 
400k steps, mixing embedding 
dimension 192, hypernetwork embedding 256 with 2 layers.

\subsection{Metrics}
\textbf{Task success rate} ($\uparrow$) measures the percentage of
episodes in which the full medical task is completed.
\textbf{JFI} ($\uparrow$) measures workload balance; values near 1
indicate equal agent contributions.
\textbf{Constraint satisfaction} (CSat, $\uparrow$) is the proportion
of evaluation episodes in which $\mathrm{JFI}(w)\ge\tau$; a value of
1.0 means the constraint was met in every episode.
The \textbf{fairness multiplier $\lambda$} indicates enforcement
pressure: high values signal active constraint violation during
training, while values near zero signal the constraint is satisfied.

\begin{table}[t]
\scriptsize
\centering
\caption{Performance comparison on the CPR task. $\uparrow$ indicates higher is better. All JFI values computed identically: end-of-episode cumulative workload JFI (excluding \texttt{noop} and \texttt{move} actions), averaged over 5 runs. $\dagger$ denotes statistical significance over fixed-penalty baselines (Mann-Whitney U, Bonferroni corrected, $p<0.0125$).}
\label{tab:results}
\begin{tabular}{lcccc}
\toprule
\textbf{Method} & \textbf{Task Success Rate ($\uparrow$)} & $\boldsymbol{\lambda}$ & \textbf{Workload JFI ($\uparrow$)} & \textbf{Constraint Sat. ($\uparrow$)} \\
\midrule
\multicolumn{5}{l}{\textit{Fixed Penalty Baselines}} \\
\textbf{QMIX} ($\lambda=0$) & $0.83 \pm 0.02$ & 0 (fixed) & $0.33 \pm 0.10$ & -- \\
\textbf{QMIX with fairness penalty} ($\lambda=10$) & $0.87 \pm 0.02$ & 10 (fixed) & $0.43 \pm 0.14$ & -- \\
\textbf{QMIX with fairness penalty}  ($\lambda=30$) & $0.90 \pm 0.03$ & 30 (fixed) & $0.47 \pm 0.12$ & -- \\
\midrule
\multicolumn{5}{l}{\textit{Adaptive Constraint Enforcement (\textbf{\texttt{AdaFair-MARL}})}} \\
\textbf{\texttt{AdaFair-MARL}} ($\tau=0.85$) & $0.86 \pm 0.04$ & $20.0 \pm 0.0$ & $0.93 \pm 0.15^\dagger$ & $0.99 \pm 0.03$ \\
\textbf{\texttt{AdaFair-MARL}} ($\tau=0.75$) & $0.90 \pm 0.06$ & $7.4 \pm 7.6$  & $0.67 \pm 0.24$         & $0.99 \pm 0.01$ \\
\textbf{\texttt{AdaFair-MARL}} ($\tau=0.65$) & $0.90 \pm 0.01$ & $5.6 \pm 2.1$  & $0.64 \pm 0.02^\dagger$ & $1.00 \pm 0.00$ \\
\textbf{\texttt{AdaFair-MARL}} ($\tau=0.55$) & $0.93 \pm 0.02$ & $3.6 \pm 3.1$  & $0.60 \pm 0.28$         & $1.00 \pm 0.00$ \\
\bottomrule
\end{tabular}
\end{table}

\begin{figure*}[t]
\centering
\includegraphics[width=0.8\linewidth]{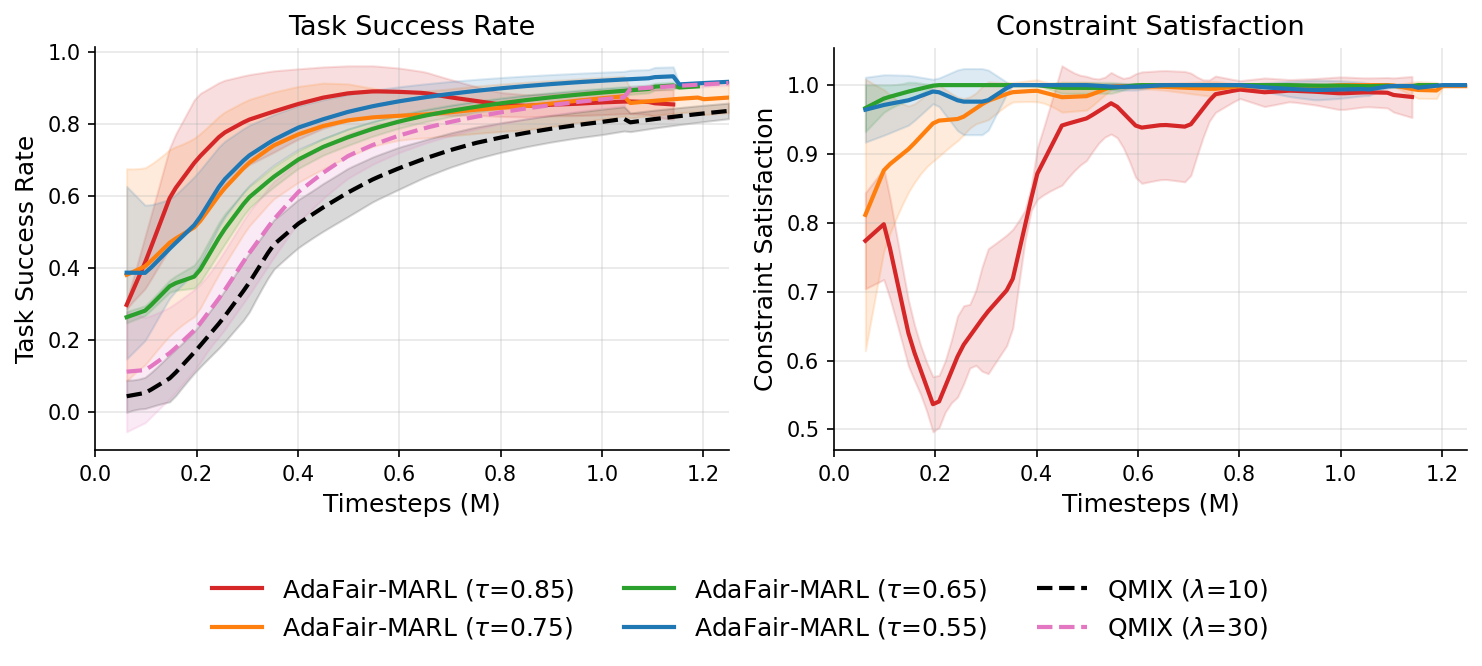}
\caption{Training curves for \textbf{\texttt{AdaFair-MARL}} variants and 
fixed-penalty baselines on the CPR task. Solid lines show 
\textbf{\texttt{AdaFair-MARL}} variants; dashed lines show fixed-penalty 
baselines. Shaded regions indicate one standard deviation across 5 runs.
\textit{Left:} Task success rate --- all methods converge to comparable task 
performance ($>$0.85), demonstrating that fairness enforcement in 
\textbf{\texttt{AdaFair-MARL}} does not degrade task efficiency relative to 
fixed-penalty baselines.
\textit{Right:} Constraint satisfaction rate --- \textbf{\texttt{AdaFair-MARL}} 
variants converge to near-perfect constraint satisfaction ($\geq$0.99) by 0.6M 
timesteps. The transient dip for $\tau=0.85$ reflects the dual ascent mechanism 
aggressively raising $\lambda$ to enforce the strictest fairness threshold before 
stabilizing. Fixed penalty baselines are absent from this panel as they 
have no constraint enforcement mechanism and thus provide no guarantee that a 
target fairness level is met.}
\label{fig:learning_curves}
\end{figure*}

\section{Results: Fairness--Efficiency Trade-offs under Varying $\tau$ Threshold}
\label{sec:results}

Table~\ref{tab:results} reports the average performance over 5 runs.
Figure~\ref{fig:learning_curves} shows training curves.
\textbf{\texttt{AdaFair-MARL}} results illustrate the effect of varying the fairness threshold $\tau$ on the balance between task success and workload equity. 
As $\tau$ increases, the fairness constraint becomes stricter, and
the dual ascent mechanism automatically raises $\lambda$ until the constraint is satisfied. 
At $\tau=0.55$, \textbf{\texttt{AdaFair-MARL}} achieves the
highest task success ($0.93 \pm 0.02$) with competitive workload balance (JFI: $0.60 \pm 0.28$); at $\tau=0.85$, stricter enforcement maintains high task success ($0.86 \pm 0.04$) while achieving the highest workload fairness (JFI: $0.93 \pm 0.15$). 
The learned multipliers adapt automatically to constraint tightness, ranging from $\lambda = 3.6 \pm 3.1$ at $\tau=0.55$
to $\lambda = 20.0 \pm 0.0$ at $\tau=0.85$, without any manual tuning.

\textbf{Fixed Penalties.}
A key distinction between \textbf{\texttt{AdaFair-MARL}} and fixed-penalty baselines is not raw JFI magnitude, but constraint satisfaction.
Fixed-penalty QMIX baselines provide no guarantee that a target fairness level
is met: regardless of $\lambda$, agents may converge to specialized roles
that maximize task reward while absorbing the static penalty, leaving
workload imbalance unresolved. Fixed-penalty baselines achieve workload JFI
of $0.43 \pm 0.14$ ($\lambda=10$) and $0.47 \pm 0.12$ ($\lambda=30$),
with no mechanism to enforce a target fairness level.
In contrast, \textbf{\texttt{AdaFair-MARL}} achieves constraint satisfaction
rates of $0.99$--$1.00$ across all $\tau$ values, confirming that the dual
ascent mechanism reliably enforces the fairness constraint. A practitioner
can set $\tau$ directly to a desired fairness level; fixed-penalty
methods offer no equivalent guarantee regardless of how $\lambda$ is chosen.

\textbf{Fairness--Efficiency Trade-off.}
The threshold $\tau$ provides explicit, principled control over the
fairness--efficiency frontier. At $\tau=0.55$, \textbf{\texttt{AdaFair-MARL}}
achieves the best task success ($0.93 \pm 0.02$) among all \textbf{\texttt{AdaFair-MARL}}
variants while satisfying the fairness constraint (CSat $= 1.00$).
At $\tau=0.65$ and $\tau=0.75$, task success remains high
($0.90 \pm 0.01$ and $0.90 \pm 0.06$ respectively) with CSat $\geq 0.99$.
At $\tau=0.85$, the penalty $\lambda$ saturates at $20.0 \pm 0.0$,
reflecting persistent constraint enforcement, yet task success
remains competitive at $0.86 \pm 0.04$. This trade-off is controllable
by design through $\tau$, whereas fixed-penalty methods (i.e., QMIX with fairness penalties\cite{siddique_fairness_nodate,ekpo_skill-aligned_2025}) provide no such
control mechanism.
\textbf{Statistical Testing.}
We evaluate statistical significance using Mann-Whitney U tests with
Bonferroni correction for multiple comparisons ($n_{\text{comparisons}}=4$,
corrected $\alpha=0.0125$). Effect sizes are reported as Cohen's $d$,
with $|d| \geq 0.8$ indicating large effects. All experiments use $n=5$ runs.
\textbf{\texttt{AdaFair-MARL}} at $\tau=0.85$ achieves significantly higher
workload JFI than fixed-penalty baselines ($p=0.001$, $d=3.62$, large effect).
\textbf{\texttt{AdaFair-MARL}} at $\tau=0.65$ similarly achieves significant
improvement ($p=0.002$, $d=1.76$, large effect). Variants $\tau=0.75$
($p=0.064$) and $\tau=0.55$ ($p=0.109$) do not survive Bonferroni correction,
though both exhibit large effect sizes ($d=1.54$ and $d=0.81$ respectively),
suggesting the non-significance is driven by within-group variance.

\section{Conclusion}
\label{sec:conclusion}

We introduced \textbf{\texttt{AdaFair-MARL}}, a constrained cooperative MARL framework whose core algorithmic component is a primal--dual update that enforces workload fairness via adaptive Lagrange multiplier updates. 
\textbf{\texttt{AdaFair-MARL}} demonstrates that equilibrium constraints can regulate 
coordination among heterogeneous agents while maintaining task 
performance, without introducing instability or reliance on manually tuned reward penalties.
Future work will extend \textbf{\texttt{AdaFair-MARL}} to settings with multiple fairness dimensions, such as skill-task alignment or multi-objective trade-offs between workload balance and performance. 
Further analysis of convergence properties under function approximation and experiments in human-AI collaboration scenarios remain important next steps.






\bibliographystyle{unsrtnat} 
\bibliography{r}

\newpage
\appendix

\end{document}